\documentclass[letterpaper, 10 pt, conference]{ieeeconf}  

\IEEEoverridecommandlockouts                              

\usepackage{graphics} 
\usepackage{epsfig} 
\usepackage{mathptmx} 
\usepackage{times} 
\usepackage{amsmath} 
\usepackage{amssymb}  
\usepackage{cleveref}
\usepackage{graphicx}
\usepackage{multirow}
\usepackage{booktabs}
\usepackage{tabularx}

\title{\LARGE \bf
Learning to Dock: A Simulation-based Study on Closing the Sim2Real Gap in Autonomous Underwater Docking
}

\author{Kevin Chang$^1$, Rakesh Vivekanadan$^1$, Noah Pragin$^1$, Sean Bullock$^1$, and Geoffrey A. Hollinger$^1$
\thanks{$^1$Collaborative Robotics and Intelligent Systems (CoRIS) Institute, \{\tt\small changk2, \tt\small vivekanr, \tt\small praginn, \tt\small bullocse, \tt\small geoff.hollinger\}\tt\small @oregonstate.edu}
\thanks{This work was funded in part by the U.S. Department of Energy under cooperative agreement DE-EE0009449.}
}

\begin{document}

\maketitle
\thispagestyle{empty}
\pagestyle{empty}

\begin{abstract}

Autonomous Underwater Vehicle (AUV) docking in dynamic and uncertain environments is a critical challenge for underwater robotics. Reinforcement learning is a promising method for developing robust controllers, but the disparity between training simulations and the real world, or the sim2real gap, often leads to a significant deterioration in performance. In this work, we perform a simulation study on reducing the sim2real gap in autonomous docking through training various controllers and then evaluating them under realistic disturbances. In particular, we focus on the real-world challenge of docking under different payloads that are potentially outside the original training distribution. We explore existing methods for improving robustness including randomization techniques and history-conditioned controllers. Our findings provide insights into mitigating the sim2real gap when training docking controllers. Furthermore, our work indicates areas of future research that may be beneficial to the marine robotics community.

\end{abstract}

\section{INTRODUCTION}

Autonomous underwater vehicles (AUVs) are a critical technology for the future of ocean exploration \cite{subseaexploration}, marine research \cite{biohotspot}, and underwater infrastructure maintenance \cite{pipelinemonitoring}. Autonomous docking, specifically, is critical for long- and short-term AUV missions, enabling data transmission, battery charging, and recovery. Typically, when docking, and AUV will attempt to land onto an object, which may connect to a wave energy converter or a device for storing or transferring data. However, operating in underwater environments poses unique challenges that traditional docking control methods struggle to address effectively. These challenges include complex, dynamic conditions, such as varying current and wave forces, limited visibility, and pressure changes, that make near-optimal navigation and control particularly challenging. Furthermore, when AUVs are deployed in the real world, different payloads such as cameras or environmental sensors are often attached in order to collect data. Traditional methods must often be re-tuned to account for these new payloads, which adds significant time and labor to a mission.

\begin{figure}[t]
\centerline{\includegraphics[width=0.5\textwidth]{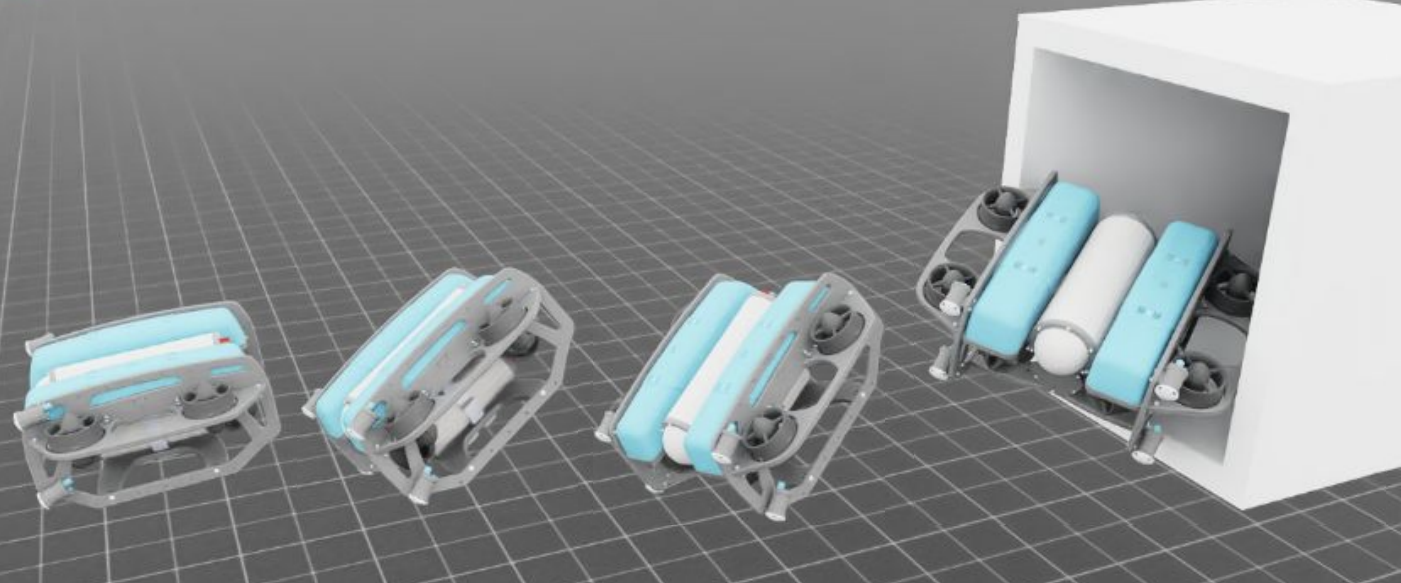}}
\caption{\footnotesize Visualization of docking trajectory in simulation environment. The vehicle starts at a random position in front of the docking station. At every iteration, the vehicle's state is passed into a trained policy which outputs thruster commands, and then those commands are passed into the simulation to update the vehicle's state. The goal is for the vehicle to land inside the docking station, minimizing its position to it and its angular distance from the forward direction.}
\label{fig:docking_ex}
\end{figure}

In addition to these challenges, the high cost and complexity of real-world testing creates significant barriers to the iterative development of AUV control systems. Reinforcement learning (RL) offers a promising approach to overcome these challenges by enabling AUVs to learn optimal control policies through experience in simulation. Further, while traditional control methods rely on explicit mathematical models, RL agents can adapt to changing environmental conditions and learn complex behaviors through trial and error. This learning-based approach is particularly valuable in underwater scenarios where accurate physical modeling is difficult due to complex fluid dynamics and environmental uncertainties. However, a common issue that arises when controlling robots with RL is deterioration in performance during real-world deployment due to disparities between simulation and the real world, also known as the sim2real gap. Common approaches to this problem include improving the simulator's realism through techniques such as system identification or learning more robust policies through dynamically randomizing environmental parameters during training \cite{sim2realquad}. Our work seeks to explore and mitigate this issue in the autonomous underwater docking domain.

In this work, we utilize an underwater robotics simulator to represent and study the sim2real problem in the context of autonomous underwater docking without performing any real-world deployments. We accomplish this by using our simulation to model the potential disturbances an AUV would encounter during deployment, then using it to evaluate the zero-shot adaptiveness of RL-based controllers. In particular, we focus our efforts on addressing the realistic scenario of docking an AUV that is weighed down by various payloads such as cameras and other tools for collecting scientific data. To develop our simulation, we build upon the underwater simulation environment of Cai et al. \cite{learningtoswim}, extending it to support generalized vehicle dynamics and rapid environment adaptation specifically for docking tasks. We also investigate how various techniques can be used to improve the robustness of RL-based controls.

\section{RELATED WORKS}

Autonomous docking of underwater vehicles presents significant challenges due to nonlinear hydrodynamics, sensor uncertainties, and environmental disturbances. Traditional docking control has relied on proportional-integral-derivative (PID) controllers, sliding mode control (SMC), and model predictive control (MPC) to navigate AUVs to docking stations \cite{anderlini, rakeshdocking}. However, these methods struggle with generalization to out-of-distribution (OOD) conditions, limiting their robustness \cite{patil}.

Recent advancements leverage deep reinforcement learning (DRL) to train policies capable of adapting to complex dynamics. Patil et al.~\cite{patil} conducted a benchmarking study comparing various state-of-the-art DRL algorithms for AUV docking. Similarly, Anderlini et al. \cite{anderlini} applied deep Q-networks (DQN) and deep deterministic policy gradients (DDPG) to docking control, showing that DQN provided smoother control signals and lower final velocities, making it more viable for real-world deployment. Furthermore, existing work shows that the reward function design plays a crucial role in RL-based docking. Patil et al.~\cite{patil} utilized a distance-based penalty as a primary reward component, ensuring that the AUV was incentivized to minimize its displacement from the docking location. Their formulation also penalized excessive thruster utilization, encouraging energy-efficient docking maneuvers. Anderlini et al. \cite{anderlini} incorporated orientation-based penalties to maintain proper alignment with the docking station, as well as state-dependent penalties to ensure a gradual reduction in velocity as the vehicle approached its goal. Despite these advancements, existing methods fail to consider the realistic scenario of autonomously docking an AUV under varying payloads.

Furthermore, significant progress has been made towards reducing the sim2real gap. One method involves dynamically randomizing environmental parameters through a process called domain randomization (DR) during training in order to learn more robust policies. This technique has shown empirical success across many types of robots such as AUVs \cite{learningtoswim}, quadrupeds \cite{learningtowalkparallel}, quadcopters \cite{multireal}, and humanoids \cite{sim2realboxlocomanipulation}. Additionally, history-conditioned policies have been shown to perform more robustly than their reactive counterparts, especially when combined with DR \cite{blindcassie, cassieremembers}.

Our work attempts to validate and build upon these studies by implementing proximity and orientation-based rewards, randomized environments, and history-based policies within the NVIDIA Isaac Sim robot simulation framework. By systematically evaluating how different techniques affect policy robustness under realistic payloads, we aim to address the sim2real problem by establishing generalizable docking controllers.

\section{METHODS}

\begin{figure}[t]
\centerline{\includegraphics[width=0.45\textwidth]{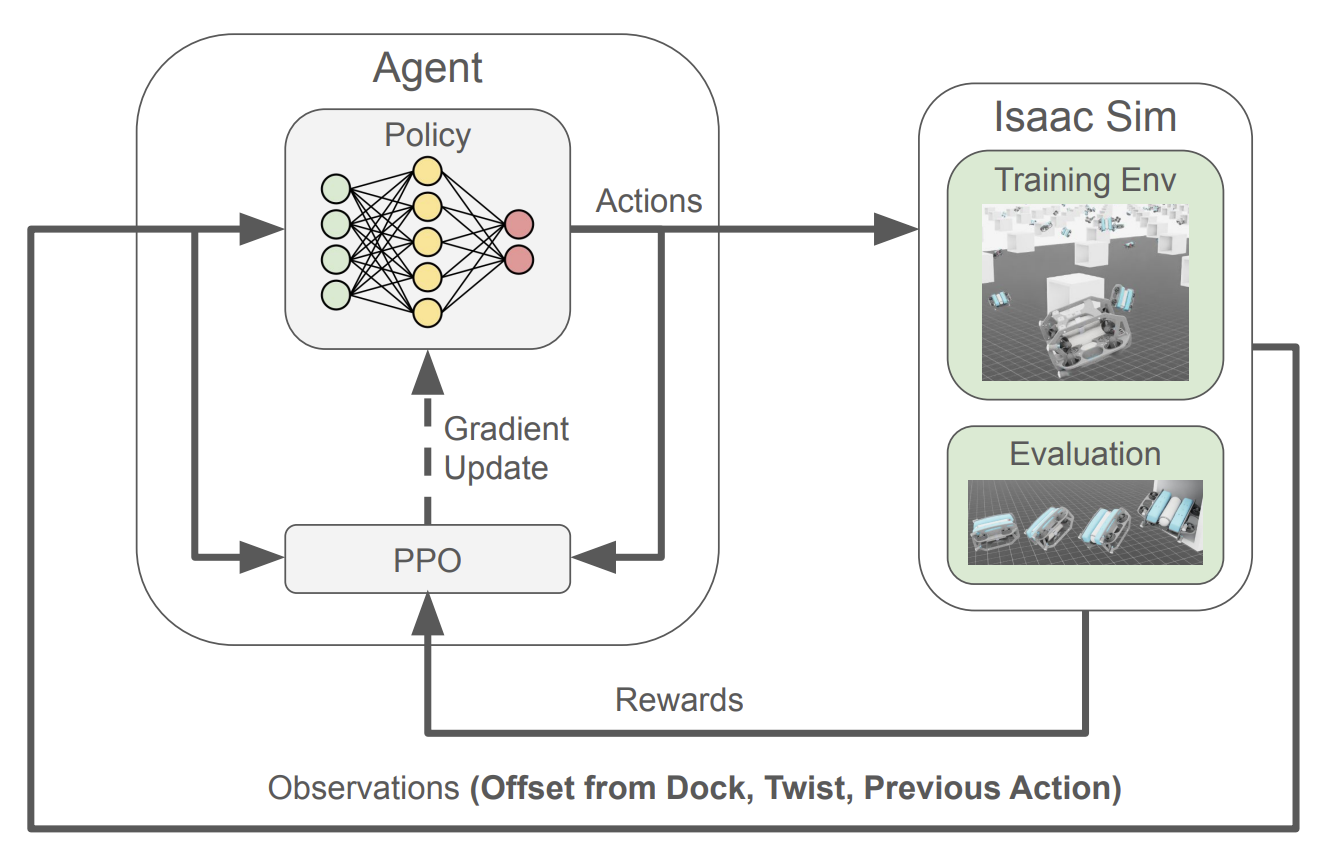}}
\caption{\footnotesize Our general flow for training and evaluating policies for autonomous docking. We utilize environments built with the Isaac Sim framework to both train and evaluate our policies.}
\label{fig:sys_diagram}
\end{figure}

We present our methodology for developing RL-based docking controllers for AUVs that are robust to varying payloads. In particular, we look to train policies in simulation and then evaluate them under simulation but with different payloads to test robustness. To standardize our results as much as possible, we specifically develop our controller for the BlueROV2 Heavy, though our methods are easily adaptable to other thruster-driven vehicles. Our proposed system is outlined in Figure \ref{fig:sys_diagram}.

\subsection{Simulation and Environment}

\subsubsection{Backend}

\begin{figure}[t]
\centerline{\includegraphics[width=0.4\textwidth]{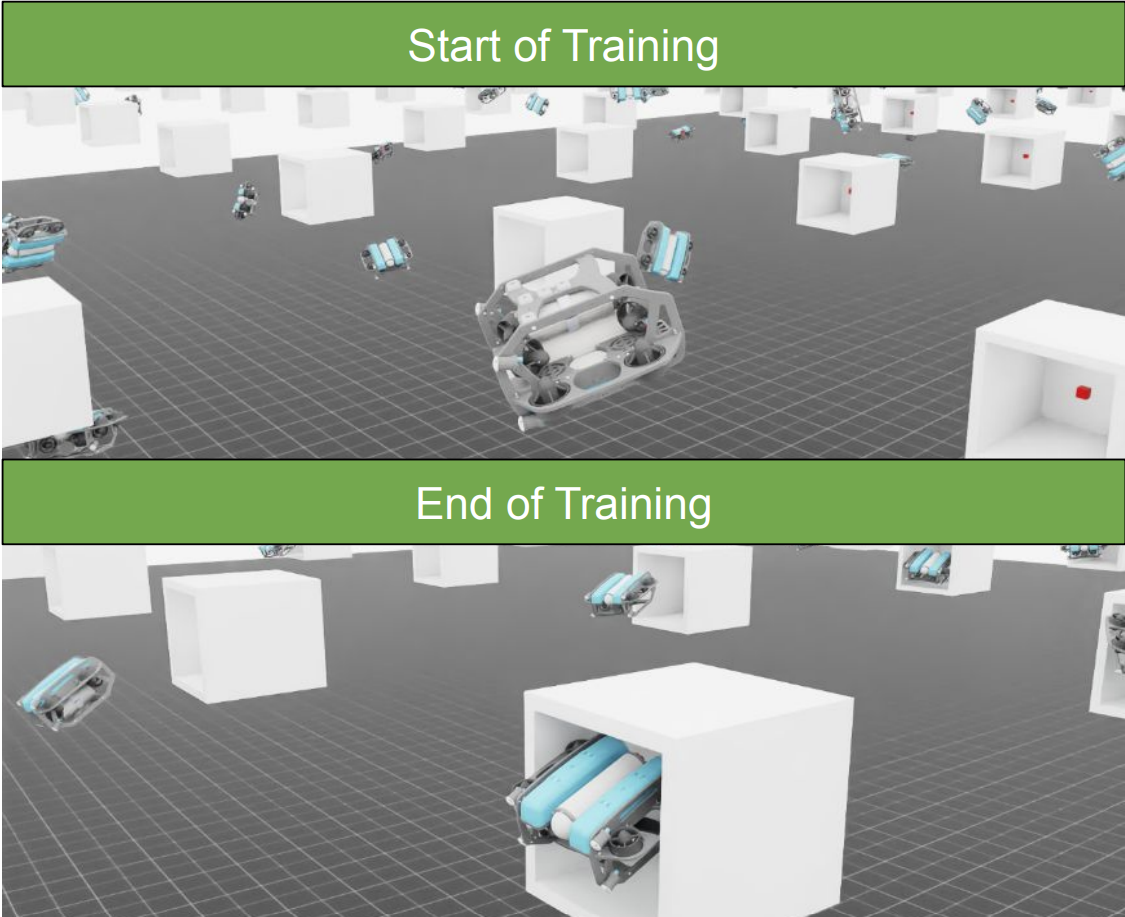}}
\caption{\footnotesize Demonstration of Isaac Sim's capability for highly-parallelized training.}
\label{fig:parallel_envs}
\end{figure}

Our methods revolve around the use of an AUV simulator built on top of NVIDIA's Isaac Labs framework, which was formerly Isaac Gym \cite{isaaclab, isaacgym}. Isaac Labs is a framework for learning-based control built on top of Isaac Sim which is itself built on top of Omniverse. Its previous versions have been leveraged to train policies for controlling quadrupeds \cite{learningtowalkparallel}, quadcopters \cite{isaacgym_drones} and AUVs \cite{learningtoswim} with demonstrated success in the real world. The key benefit of Isaac Labs is that as seen in Figure \ref{fig:parallel_envs}, it enables simulating thousands of robots in parallel with very little GPU usage, enabling fast learning of complex and diverse behaviors. Additionally, the intuitive API strikes an ideal balance between robustness and flexibility, enabling new environments to be implemented fairly trivially while simultaneously supporting custom components such as alternate dynamics models. Furthermore, the framework utilizes the Universal Scene Description (USD) format, making it straight-forward to design detailed 3-dimensional scenes. An AUV simulator was previously built using this framework by Cai et al. \cite{learningtoswim}, so in this work we utilize their existing code and extend it to support our research. In particuular, we generalize the software to support rapid adaptability between different AUVs, such as CUREE \cite{curee} and BlueROV2 Heavy, by abstracting hydrodynamic modeling and providing a modular environment setup. This enables streamlined integration of realistic payload dynamics, docking stations, and enhanced state representations.

\subsubsection{AUV Dynamics Model}

We replicate how Cai et al. \cite{learningtoswim} models the vehicle dynamics but slightly adapt it to support the BlueROV2 Heavy. For completeness, we describe the underlying physics model here. Non-hydrodynamic effects and collisions are handled by the PhysX engine, and then hydrodynamics are modeled based on MuJoCo's simple inertial box model \cite{mujoco}. For simplicity, thrusters are assumed to have no latency and are modeled using a zero-order dynamics model. Then, the angular velocities of the thrusters are converted to linear forces using the model proposed by Yoerger et al. \cite{thrusterdynamics}.

\subsubsection{Docking Environment}

For the docking task, we design an environment that realistically models the real-world problem. For the docking station itself, we use a $0.7~m \times 0.7~m \times 0.7~m$ hollow cube with one face open. Additionally, to ensure that the AUV can dock from a diverse array of starting positions, we randomly sample the starting position from a $2~m \times 2~m \times 2~m$ cube in front of the open face of the docking station. We do not consider the case where the AUV must start behind the docking station because if we were to deploy the policy in the real world, the docking station would be localized using an AprilTag pointing outwards, which would only be visible from the front. The environment along with one of our generated trajectories is shown in Figure \ref{fig:docking_ex}.

\subsection{Training}

\subsubsection{Algorithm}

To train our AUV docking policies, we employ PPO \cite{ppo}, an on-policy, policy gradient algorithm that offers a favorable balance between sample efficiency, implementation simplicity, and performance. Our decision to use PPO was informed by the vast amount of existing work in RL for robotic control that has demonstrated success with the algorithm across many domains \cite{animalstyle}, \cite{learningtoswim, rlildrone}, \cite{sim2realboxlocomanipulation}. Additionally, given that we are simulating thousands of robots in parallel, PPO works very well because as an on-policy, policy gradient method, the large batch size will lead to a more stable, informed, and accurate gradient. On the other hand, state-of-the-art off-policy algorithms such as SAC will likely struggle to handle the high number of parallel environments. Our use of PPO is outlined in Figure \ref{fig:sys_diagram}.

\subsubsection{Actions and Observations}

Much of our action and observation space is built off of the existing work by Cai et al. \cite{learningtoswim}. Since we are developing controls for BlueROV2 Heavy, our action space will consist of $8$-dimensional vectors with each value correlating with a PWM value in each of the $8$ thrusters. We define a single observation at timestep $t$ as $\vec{o}_t = (x_t, q_t, \dot{x}_t, \dot{q}_t, a_t)\in S$ where $x_t\in \mathbb{R}^3$ is the relative position of the docking station from the robot, $q_t\in \mathbb{R}^4$ is orientation of the AUV in quaternion form, $\dot{x}_t \in \mathbb{R}^3$ is the AUV's linear velocities, $\dot{q}_t \in \mathbb{R}^3$ is the AUV's angular velocities in Euler form, and $a_t\in A$ is the actions taken in the previous iteration, all at timestep $t$. Then, we also incorporate history into our observation space. Our history-augmented observation becomes $\vec{\mathcal{O}}_t = (o_{t-h+1}, o_{t-h+2}, \ldots, o_{t-1}, o_t)$ where $h$ is the history length and represents the number of observations that are concatenated.

\subsubsection{Reward Function Formulation}

For our reward function, we determine a simple reward function that achieves our desired behavior:

\begin{equation}
\begin{split}
R_t &= \lambda_1 R_{\text{dist}} + \lambda_2 R_{\text{orient}} \\
\end{split}
\end{equation}

where each term is defined as:

\begin{align}
R_{\text{dist}} &= \exp(-||p_{t} - p_{\text{dock}}||_2) \\
R_{\text{orient}} &= \exp(-||\theta_{t} - \theta_{\text{dock}}||_2) \\
\end{align}

where:
\begin{itemize}
    \item $p_t$, $\theta_t$ are the AUV's position and orientation at time $t$
    \item $p_{\text{dock}}$ and $\theta_{\text{dock}}$ are the target docking position and orientation
\end{itemize}

The weighting coefficients $\{\lambda_1 = 0.2, \lambda_2 = 0.03\}$ control the relative importance of each reward component. To determine the best values for these coefficients, we started by training with $\lambda_2 = 0$, and then gradually increased it until we attained more stable trajectories while still offering the AUV flexibility to tilt in order to point its thrusters more efficiently.

\subsubsection{Domain Randomization and State History}

We implement DR during training to enhance robustness against varying payloads. Our approach focuses specifically on randomly generating different payloads during training.

The DR process consists of generating a payload with mass uniformly sampled from $0$ to an upper bound and then spawning it at a position around the AUV uniformly sampled from a sphere with a specified radius. This procedure creates variability in both the magnitude and distribution of mass affecting the vehicle dynamics.

This approach simulates manufacturing variations, equipment reconfigurations, and payload changes that an AUV might experience in real-world deployments. By training across this distribution of dynamics, we aim to develop policies that are robust across varying payloads.

Additionally, we compare policies trained using only the latest state observation against those concatenated with a history of previous states. We hypothesize that incorporating history will allow our policy to learn behaviors that are more specific to different hydrodynamic conditions.

\subsubsection{Training Process}

All policies are trained in NVIDIA Isaac Sim using parallel environments to accelerate data collection. Again following the existing work by Cai et al. \cite{learningtoswim}, we simulate 2048 AUVs in parallel. Through empirical tests we found that $500$ iterations was normally enough to reach convergence, so we allow all policies to train for that many iterations. Policy networks employ a multi-layer perceptron (MLP) architecture.

\subsection{Evaluation Methods}

To evaluate the robustness of the trained docking policies, we assess their performance in simulation under a variety of payloads. Ideally, we would also evaluate our policies in the real world, but due to time constraints we were limited to simulation-only. However, simulation does allow us to have significantly more control over the hydrodynamics seen during evaluation and so we believe that it is an effective way to study the robustness of our proposed policies.

\section{EXPERIMENTS}

\subsection{Policy Architectures and Training Approaches}

We implement and evaluate four distinct policy architectures to investigate their performance in AUV docking tasks. The configurations for these policies are shown in Table \ref{tab:training_configs}. All policies are implemented as multi-layer perceptrons (MLPs) with identical hidden layer structures to ensure a fair comparison, with only the input dimension differing between architectures. Each configuration is trained over $3$ seeds for reproducibility.

\begin{table}[h]
\centering
\fontsize{10}{12}\selectfont
\resizebox{\columnwidth}{!}{
\begin{tabular}{|l|cc|c|}
\hline
\textbf{Policy} & \multicolumn{2}{c|}{\textbf{Payload}} & \textbf{\textbf{History Length}} \\
\cline{2-3}
& \textbf{Mass (kg)} & \textbf{Spawn Radius (m)} & \\
\hline
Naive & None & None & 1 \\
Small DR & $[0, 2.5]$ & $0.1$ & $1$ \\
Large DR & $[0, 5.0]$ & $0.3$ & $1$ \\
Large DR w/ History & $[0, 5.0]$ & $0.3$ & $3$ \\
\hline
\end{tabular}
}
\caption{Training configurations for docking controllers.}
\label{tab:training_configs}
\end{table}

\subsection{Evaluation Protocol}

We evaluate all four policy configurations under several payload settings, shown in Table \ref{tab:eval_scenarios}. Each policy undergoes 20 evaluation episodes per condition, with randomized initial positions within a designated starting zone. The docking station remains in a fixed position throughout all evaluations.

\begin{table}[h]
\centering
\resizebox{0.8\columnwidth}{!}{%
\begin{tabular}{|l|c|c|}
\hline
\textbf{Scenario} & \textbf{Payload Mass (kg)} & \textbf{Payload Position (m)} \\
\hline
Easy & None & None \\
Medium & $3.5$ & $0.15$ (x-axis) \\
Hard & $7.0$ & $0.3$ (x-axis) \\
\hline
\end{tabular}%
}
\caption{Payload configurations for evaluating policies under different levels of real-world disturbances.}
\label{tab:eval_scenarios}
\end{table}

\subsection{Performance Metrics}

We employ multiple metrics to assess policy performance comprehensively. We analyze both the AUV's euclidean distance to the docking station and its angular distance from the forward direction with respect to time to study the smoothness and speed at which different policies dock the AUV.

\section{RESULTS AND DISCUSSION}

\subsection{Training}

In studying the training curves of each policy, we see that most policies converged by around 200 iterations. There does not seem to be any major difference in training speed across the different training configurations.

\subsection{In-Sim Evaluation}

\begin{figure}[htbp]
\centerline{\includegraphics[width=0.48\textwidth]{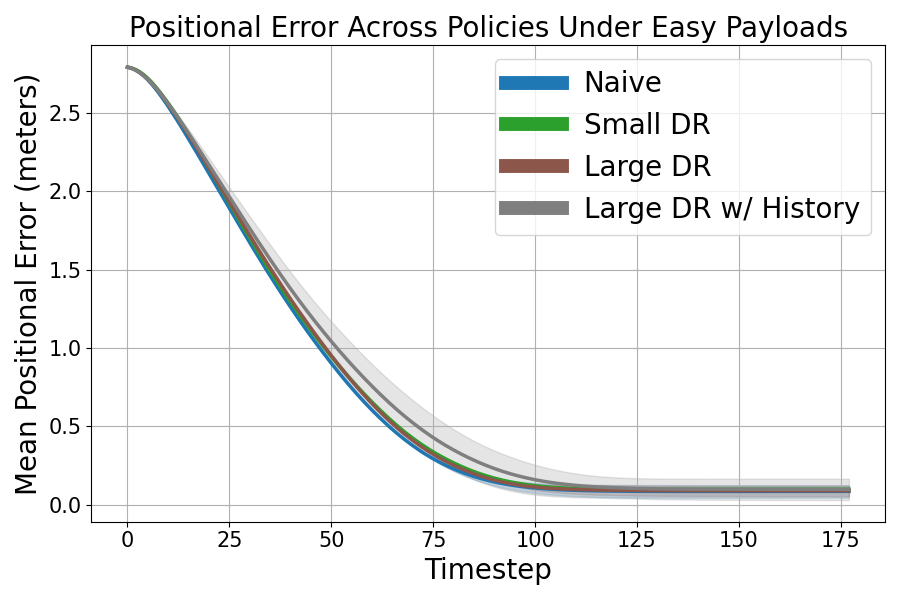}}
\caption{\footnotesize Positional error over time under the easy payload configuration.}
\label{fig:pos_easy}
\end{figure}

\begin{figure}[htbp]
\centerline{\includegraphics[width=0.48\textwidth]{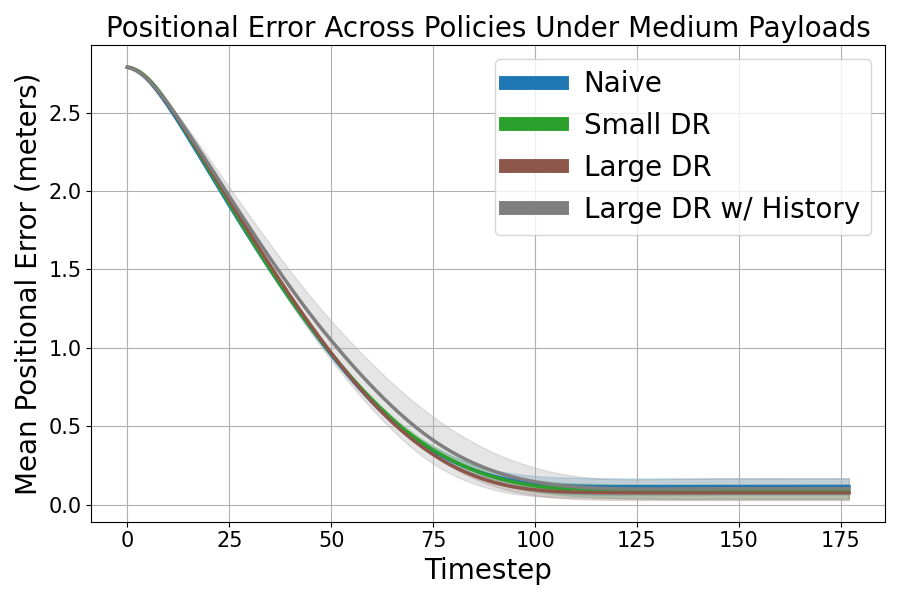}}
\caption{\footnotesize Positional error over time under the medium payload configuration.}
\label{fig:pos_medium}
\end{figure}

\begin{figure}[htbp]
\centerline{\includegraphics[width=0.48\textwidth]{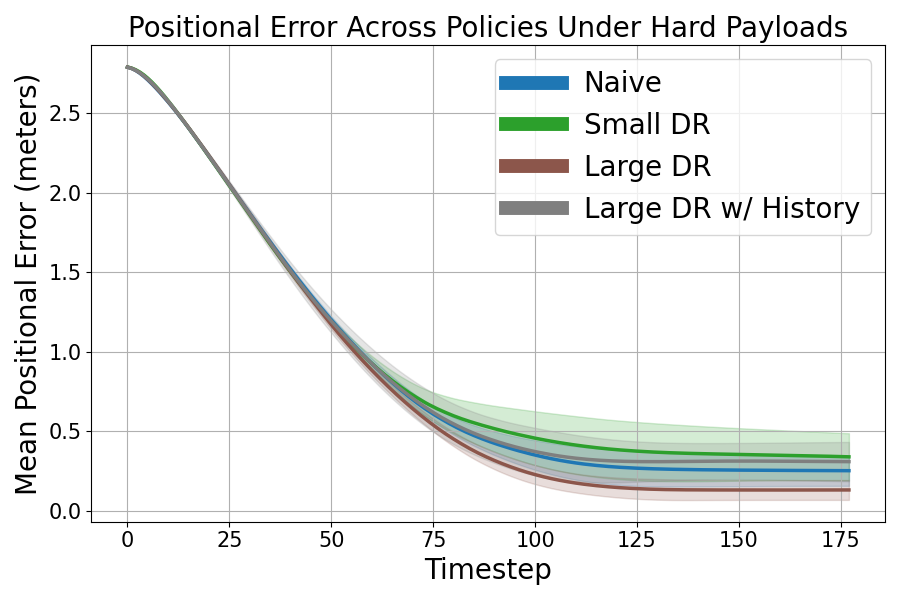}}
\caption{\footnotesize Positional error over time under the hard payload configuration.}
\label{fig:pos_hard}
\end{figure}

\textbf{Positional Error:} We find that generally, all policies perform roughly equally well regardless of DR or history length on easy to medium level payloads, as can be seen in Figures \ref{fig:pos_easy}, \ref{fig:pos_medium}. Interestingly, we note that the history-based policies exhibit higher variance, possibly due to overfitting. Under hard payloads, we observe that the policies trained with a large amount of DR performs best, as can be seen in Figure \ref{fig:pos_hard}. We hypothesize that the similarity in performance across all training configurations is due to how attaching varying payloads does not significantly change the actions that a policy should optimally output given some input state. Notably, we find that when we incorporate state history, which should lead to more specified behaviors, the AUV's performance under extreme scenarios is only as strong as the naively trained policies.

\begin{figure}[htbp]
\centerline{\includegraphics[width=0.48\textwidth]{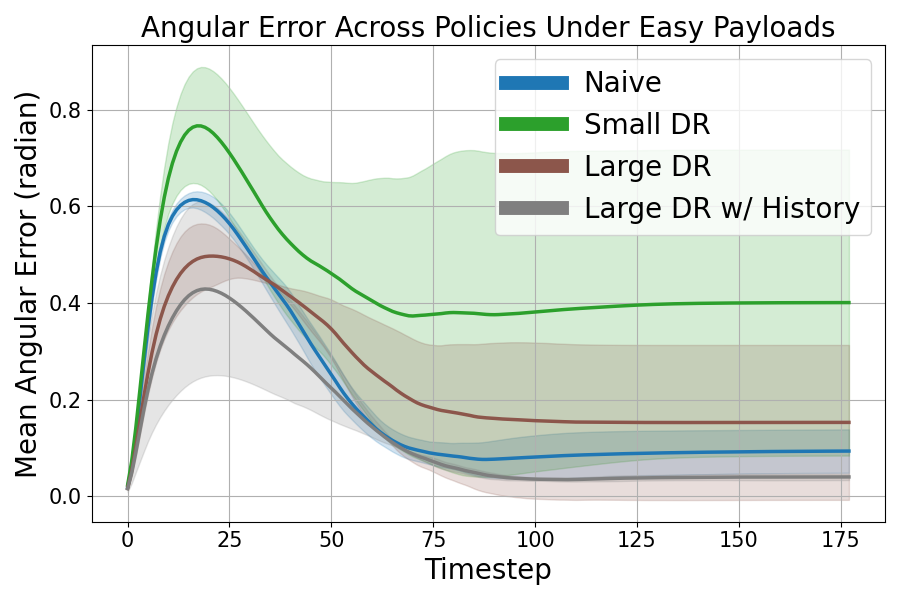}}
\caption{\footnotesize Angular error over time under the easy payload configuration.}
\label{fig:ang_easy}
\end{figure}

\begin{figure}[htbp]
\centerline{\includegraphics[width=0.48\textwidth]{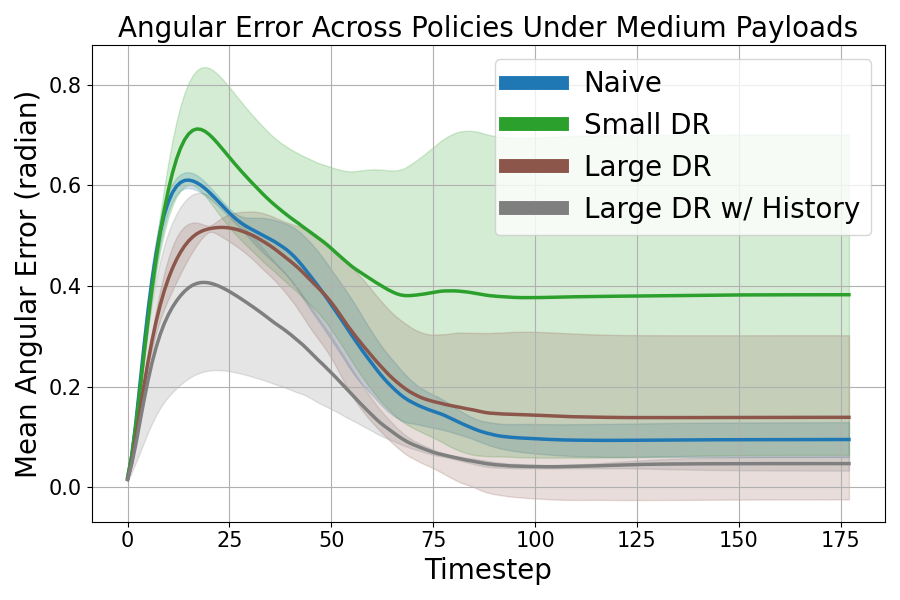}}
\caption{\footnotesize Angular error over time under the medium payload configuration.}
\label{fig:ang_medium}
\end{figure}

\begin{figure}[htbp]
\centerline{\includegraphics[width=0.48\textwidth]{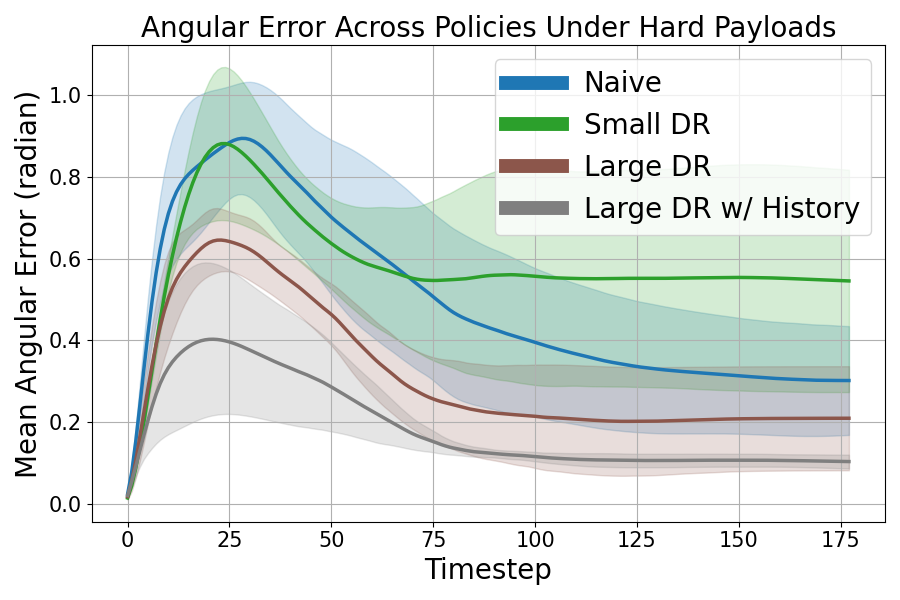}}
\caption{\footnotesize Angular error over time under the hard payload configuration.}
\label{fig:ang_hard}
\end{figure}

\textbf{Angular Error:} With regards to angular error, we see fairly significant differences between different policies, as can be seen in Figures \ref{fig:ang_easy}, \ref{fig:ang_medium}, \ref{fig:ang_hard}. We do observe across all payload settings, the policies trained using a smaller amount of DR consistently perform the worst. When analyzed qualitatively, we observed that these policies tended to primarily be misaligned along the Z-axis of rotation. This suggests that it is possible that increasing the weight of the orientation term of the reward function would significantly improve this error metric. Another fascinating result is how training with a large amount of DR consistently performs slightly worse than the naively trained policies, but policies using the same amount of DR but with state history are able to marginally exceed the naive performance. This makes theoretical sense, as DR should slightly worsen performance by forcing the policy to be robust across a wider range of conditions, but incorporating history should enable the policy to condition its behavior on different environmental variations, mitigating the loss in specificity and performance.

\section{CONCLUSION}

In this work, we presented and evaluated various control policies for AUV docking, focusing on the performance implications of DR and the inclusion of historical data across a range of operational payload conditions with a focus on reducing the sim2real gap. Despite being entirely simulation-based, our study's findings offer significant insights into the design and optimization of AUV docking strategies that can maintain performance under real-world disturbances.

\textbf{Key Findings:}
\begin{itemize}
    \item \textbf{Naive Policy:} Highly effective under standard to extreme payloads, underscoring the viability of simple control mechanisms for typical scenarios.
    \item \textbf{Domain Randomization Policies:} While maintaining similar or slightly worse performance to the naively trained policy under lighter payloads, using DR demonstrates marginal improvements in extreme payload scenarios. 
    \item \textbf{History-based Policies:} Combining DR with history-based architectures marginally improves performance compared with policies trained with DR alone. It occasionally exhibits higher variance, possibly due to overfitting.
\end{itemize}

In conclusion, we find that a naively trained policy is effective for the docking task under varying, realistic payloads, though incorporating other techniques like DR and history-based architectures can marginally improve performance under extreme payloads. We note that when high accuracy is required for a successful dock, then the marginal improvements provided by these robust training techniques may be valuable. Towards reducing the sim2real gap, our results indicate that given pose and twist estimates, a policy does not require DR or memory to adapt to unexpected payloads which shift the vehicle's mass distribution. This suggests that when testing a learned docking controller in the real world, it may be most reasonable to start with a naively trained policy, and then utilize DR and memory-based architectures conservatively as real-world challenges arise.

In the future, we plan to continue exploring how DR and state history contribute to a policy's performance. Additionally, we plan to compare our learning-based method with optimization-based methods in order to better understand how RL compares with state-of-the-art controllers. Lastly, we plan to model more realistic disturbances into our simulation and eventually deploy our controller in the real world to validate and refine our simulated findings.




\section*{ACKNOWLEDGMENTS}

Thank you to Levi Cai for the helpful feedback and discussions. Also thank you to the maintainers and contributors of Isaac Sim and Isaac Labs for continuously improving the framework and creating extensive openly-available documentation.





{\small
\bibliographystyle{IEEEtran}
\bibliography{refs}
}

\end{document}